\documentclass[conference]{IEEEtran}

\usepackage{setspace,tabularx}
\usepackage[tight,footnotesize]{subfigure}
\usepackage{times}
\usepackage{algorithm}
\usepackage[noend]{algorithmic}
\usepackage{amsfonts}
\usepackage{amsmath}
\usepackage{amssymb}
\usepackage{array}
\usepackage{boxedminipage}
\usepackage{caption}
\usepackage{cite}
\usepackage{color}
\usepackage{floatflt}
\usepackage{float}
\usepackage{graphics}
\usepackage{graphicx}
\usepackage{indentfirst}
\usepackage{latexsym}
\usepackage{hyperref}
\usepackage{multirow}
\usepackage{picinpar}
\usepackage{placeins}
\usepackage{psfrag}
\usepackage{subfigure}
\usepackage{theorem}
\usepackage{wrapfig}
\theoremstyle{break}

\newcommand{\ignore}[1]{ }

\newcommand{\figlbl}[1]{\label{fig.{#1}}}
\newcommand{\figref}[1]{Fig.~\ref{fig.{#1}}}

\newcommand{\beq}{\begin{equation}}
\newcommand{\eeq}{\end{equation}}

\graphicspath{{../fig/}}
\begin{document}

\title{Hardware Trojan Attacks on Neural Networks}
\author{\IEEEauthorblockN{Joseph Clements and Yingjie Lao}
\IEEEauthorblockA{Department of Electrical and Computer Engineering \\
Clemson University, Clemson, SC 29634\\
Email: jfcleme@g.clemson.edu, ylao@clemson.edu
}}
\maketitle

\begin{abstract}
With the rising popularity of machine learning and the ever increasing demand for computational power, there is a growing need for hardware optimized implementations of neural networks and other machine learning models. As the technology evolves, it is also plausible that machine learning or artificial intelligence will soon become consumer electronic products and military equipment, in the form of well-trained models. Unfortunately, the modern fabless business model of manufacturing hardware, while economic, leads to deficiencies in security through the supply chain. In this paper, we illuminate these security issues by introducing hardware Trojan attacks on neural networks, expanding the current taxonomy of neural network security to incorporate attacks of this nature. To aid in this, we develop a novel framework for inserting malicious hardware Trojans in the implementation of a neural network classifier. We evaluate the capabilities of the adversary in this setting by implementing the attack algorithm on convolutional neural networks while controlling a variety of parameters available to the adversary. Our experimental results show that the proposed algorithm could effectively classify a selected input trigger as a specified class on the MNIST dataset by injecting hardware Trojans into $0.03\%$, on average, of neurons in the 5th hidden layer of arbitrary 7-layer convolutional neural networks, while undetectable under the test data. Finally, we discuss the potential defenses to protect neural networks against hardware Trojan attacks.

\end{abstract}

\section{Introduction}
\label{sec:intro}
%Recent advances in computing power have made the implementation of neural networks and therefore machine learning much more accessible in recent years.
%\textbf{rework references}

The rapid evolution of machine learning has advanced numerous research fields and industries, including safety-critical areas such as biometric security, autonomous driving, cybersecurity, health, and financial planning~\cite{parkhi2015deep, karu1996fingerprint, chen2015deepdriving, saxe2015deep, esteva2017dermatologist, choi2016doctor, moghaddam2016stock}. Technology and human life become increasingly intertwined, which has resulted in a growing priority with regards to the security of machine learning. However, due to the ubiquity and complexity of machine learning especially deep neural networks, it has been shown recently that these techniques are quite vulnerable to well-crafted attacks~\cite{barreno2006can, papernot2016limitations, szegedy2013intriguing, mozaffari2015systematic}, which raised security concerns in the practical deployment of machine learning technologies. %For instance, machine learning algorithms can be easily fooled by adversarial samples~\cite{szegedy2013intriguing} or be poisoned by malicious training data~\cite{mozaffari2015systematic}. In other words, machine learning itself may prove the weakest point in a system where security is a concern.

%Adversarial examples\cite{szegedy2013intriguing}, poisoning attacks\cite{chen2017targeted}, and software Trojans\cite{liu2017neural,liu2017trojaning} are just a sample of the security issues being explored in the research.

%The current trend in machine learning development is to build and train a neural network model in software, running those models on CPUs or, more preferable, GPUs. This trend has understandably lead to the bias of machine learning security towards the software perspective.
Meanwhile, as the amount of available data is vastly increasing and applications are becoming tremendously sophisticated, deep learning has emerged as a promising research area that could approach human-level performance. Deep learning usually involves much larger and deeper networks, whose efficiency on large datasets becomes a bottleneck. In recent years, various specific hardware acceleration techniques have been investigated to overcome the physical constraints of certain machine learning applications \cite{ovtcharov2015accelerating, chen2014dadiannao}. Given this evolution, it is highly plausible that machine learning, in the form of well-trained models, will soon become pervasive in consumer electronic products and military equipment. However, along with opportunities, this paradigm shift also brings new security risks and concerns.

%Attacks on machine learning models in prior works can be mainly classified as either causative attacks that influence the training to obtain the desired result or exploratory attacks that seek vulnerabilities at test time or execution stages~\cite{papernot2016towards}.

Attacks on machine learning models in prior works can be mainly classified into those conducted in the training phase, controlling the training algorithm, and those in the application phase, taking advantage of faults in the well-trained model~\cite{papernot2016towards}. However, to the best of our knowledge, the analysis of the supply chain security has never been the subject of any study for adversarial machine learning. Indeed, the assumption that hardware is trustworthy and that security effort needs only encompass networks and software is no longer valid given the current semiconductor industry trends that involve design outsourcing and fabrication globalization. To expand studies on the hardware attack space on neural networks, we examine a new threat model where the adversary attempts to maliciously modify the machine learning hardware device by inserting a stealthy hardware "backdoor", i.e., hardware Trojan~\cite{chakraborty2009hardware, tehranipoor2010survey}. Through the hardware Trojan, the adversary will be able to gain access to the well-trained model or alter the prediction of the machine learning system, which could provide the adversary a strong advantage after the devices are deployed in applications. For example, an adversary in a position to profit from excessive or improper sale of specific pharmaceutics could inject hardware Trojans on a device for diagnosing patients using neural network models. The attacker could cause the device to misdiagnose selected patients to gain additional profit.

 %other option: a fingerprint scanner could be utilized to protect a device or place from unlawful access. If an attacker had inserted  a hardware trojan in this device before its implementation, this adversary could force the device to authorize his entry into the system or place as legitimate, gaining unrestricted access through the compromised device.

 %from other paper: performing hardware Trojan attacks against face recognition systems enables the attacker to impersonate another person, thus the attacker can mislead the authentication system into identifying him as a person that has access to a building or a device, so that the attacker can get into a place or a system that he originally cannot access.
\begin{table*}[t]
\centering
\small
\caption{List of notations used in this paper.}
\label{table:notation}

\begin{tabular}[b]{|c|c|p{12.8cm}|}
\hline
Notation & Name & Description \\
\hline
$F(\cdot)$  & network model & A function representing the mapping between the input and the output of a neural network. \\
\hline
$F_{l_1:l_2}(\cdot)$ & sub-network & A function implemented by a subset (i.e., from layer $l_1$ to layer $l_2$) of the neural network. \\
\hline
$f_l(\cdot)$ & activation function & The activation function of the entire layer $l$. \\
\hline
$W_l$ & weight matrix & A matrix of the weights including bias associated with layer $l$. \\
\hline
$H_l$ & intermediate value & The internal activations after hidden layer $l$ of the neural network.\\
\hline
$(X,Y)$ & labeled data & A set of input vectors $X$ and the correct labels $Y$; $x$ and $y$ denote single elements from these sets. \\
\hline
$\tilde{(\cdot)}$ & Trojan-injected instance & An element of a neural network which is compromised by a hardware Trojan.   \\
\hline
$\tilde{x}$ & input trigger & The input that triggers the malicious behavior of an injected hardware Trojan.\\
\hline
$\tilde{y}$ & intended prediction & The malicious output of the neural network when hardware Trojan is triggered.\\
\hline
${H}_{l-1}$ & Trojan trigger & The intermediate value triggers the Trojan in layer $l$. \\
\hline
$p$ & perturbation & The difference between Trojan-injected $\tilde{H_l}$ and the original $H_l$.  \\
\hline
$T$ & dynamic range  & The dynamic range of $h_l$ determines the perturbation constraint. \\
%$Y_p$ & correct output & The predictions generated by a Trojan-free network with the test set $X$ as the input. \\
%\hline
%$Y_{\sigma}$ & signature & The predictions generated by a potentially Trojan-injected model. \\
%\hline
%${\alpha}$ & stealthiness & The probability of exhibiting malicious behavior under test data. \\
\hline

\end{tabular}
\end{table*}%

In this paper, we develop a novel framework of hardware Trojan attacks on neural networks. The major contributions of this paper are summarized below:
\begin{itemize}
  \item This work introduces, for the first time, hardware Trojan attacks in the scope of neural networks. To the best of our knowledge, the only other attack on a neural network in the hardware domain comes in the form of fault injection techniques on the parameters of a well-trained network\cite{liu2017fault}. Our attack deviates from this work, as we attempt to specifically target the hardware circuitry of the network without modifying any weights. In addition, the hardware Trojan is inserted during the supply chain, while the fault injection is applied in the application phase.
  \item This paper expands the current taxonomy of neural network security to include this new attack, which also provides a basis for categorizing potential future new attacks in the hardware domain.
  \item We propose several algorithms for strategically inserting hardware Trojans into a neural network under different adversarial scenarios. %These algorithms utilize modified versions of adversarial sampling attacks to insert hardware Trojans on the internal layers of a neural network.
  \item We also discuss several possible defense approaches against the proposed attacks. %Many techniques have arisen from protection techniques against adversarial sampling for neural networks as well as hardware Trojan attacks. The effectiveness and limitations of these defense methods are discussed.
\end{itemize}

The remainder of the paper is organized as follows: \hyperref[sec:backg]{Section II} briefly reviews the basics of neural networks and hardware Trojans. \hyperref[sec:tax]{Section III} expands the current taxonomy of neural network security to more easily encompass the types of attacks possible in the hardware and then defines a threat model for our adversarial setting. In \hyperref[sec:over]{Section IV}, several novel algorithms for injecting Trojans are proposed. Then, we present the experimental setup used to test the algorithms and evaluate their performance in \hyperref[sec:exp]{Section V}. \hyperref[sec:def]{Section VI} discusses the possible defenses available to the designer against malicious hardware Trojan attacks. Finally, \hyperref[sec:con]{Section VII} presents remarks and concludes this paper.

\section{ Background }\label{sec:backg}

%In this section, we build a preliminary background for neural networks. Then, we present well-known techniques for generation and defense of adversarial examples. Finally, we introduce the basics of hardware Trojans.

\subsection{Neural Network}

A neural network is a model of machine learning that utilizes successive layers of neurons to compute an approximate mapping from an input space to an output space~\cite{goodfellow2016deep}. The input layer is usually used as a placeholder for the primary inputs, which communicates to the output layer through one or more hidden layers. Each neuron computes a weighted sum of the output vector from the previous layer, then optionally applies an activation function, and finally outputs the value, as shown in~\figref{Neuron}.

%For each layer, neurons are organized in parallel on the same inputs producing a vector of outputs. %simply passing the vector to the hidden layers. Each hidden layer transforms the output of the previous layer, passing the computed vector to the following layer. The output vector of the final layer is the primary output of the network, generalizing to a classifier this output is a probability distribution representing the likelihood that the input is from a specific class.

\begin{figure}[htbp]
\centering
\resizebox{0.3\textwidth}{!}{%
\includegraphics[]{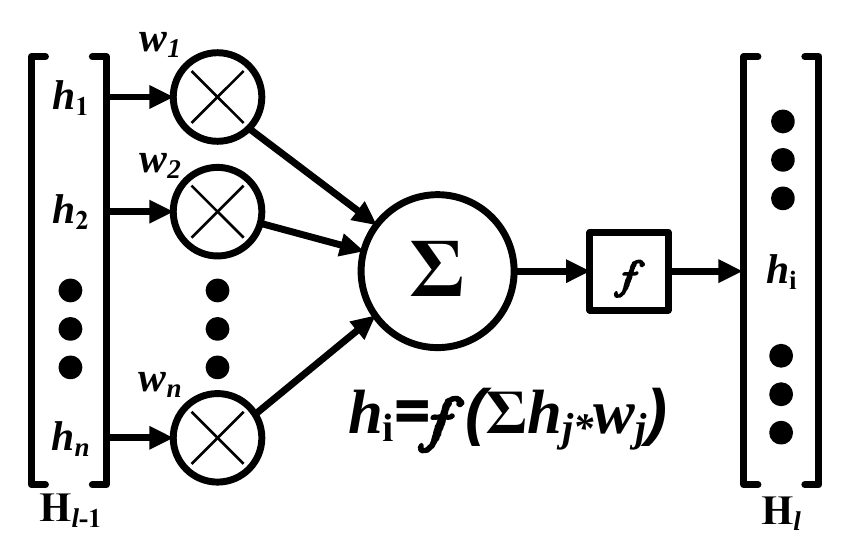}}
\caption{\small The basic operation and function of a neuron. }
\figlbl{Neuron}
%\vspace{-1em}
\end{figure}

The notations used throughout this paper are summarized in \hyperref[table:notation]{Table I}. By using these notations, the function of each layer can be formally defined as
\begin{equation}\label{eq:1}
\small H_{l}=f_{l}(W_l\cdot H_{l-1}).
\end{equation}
Consequently, by feeding the output of each layer to the input of the subsequent layer, the entire network can be characterized as: $y_{p} = F(x,W)$.

Before the neural network can produce the desired results, the parameters of the network must be trained. Utilizing a cost function, $C(y,F(x,W))$, a measure quantifying the error between the network's predicted output and its desired value under a given input, $W$ can be modified to produce the desired output. Specifically, supervised learning backpropagates the gradients of the cost function with Equation \ref{eq:BackProp} and updating the network weight iteratively.
%Supervised learning usually backpropagates the gradient of the cost function, $C(y_{label},F(x,W))$, a measure quantifying error between the networks predicted value and its desired value which can be given by %The cost function used during training is highly dependent on the network's application, in the case of classifiers a softmax function in combination with cross entropy loss are typically used. The gradient of the cost function is defined as:
\begin{equation}\label{eq:BackProp}
\small \nabla C(y,F(x,W)) = \left[ \dfrac {\delta C(y,F(x,W))}{\delta w} \right]_{w \in W}.
\end{equation}

%The first step in finding the gradient is to backpropagate the error of the cost function through the network with respect to the weights:
%\begin{equation}
%\small
%\dfrac{\delta C(y_{label},F(x,W))}{\delta W} = \dfrac{\delta C(y_{label},F(x,W))}{\delta y_L} \dfrac{\delta y_L}{\delta W}.
%\end{equation}
%By continuing the chain rule, the error can be computed and backpropagated layer by layer, which can be expressed by
%\begin{equation}
%\small
%\dfrac{\delta y_l}{\delta W} = f_l'(W_l \cdot y_{l-1}) \cdot W_l \cdot \dfrac{\delta y_{l-1}}{\delta W}.
%\end{equation}
%Finally, the gradient every $w$ in the network can be updated as $w = w - \dfrac {\delta C(y_{label},F(x,W))}{\delta w}$. The backpropagation of derivatives through neural networks has been explored in many training methods as well as many attacks on neural networks.

\subsection{Adversarial Example}
%Generalizing to neural network classifiers,
An adversarial example is an attack on a machine learning model which attempts to generate an input in such a way that it would be correctly classified by a human observer but is misclassified by the neural network~\cite{szegedy2013intriguing, akhtar2018threat, moosavi2017universal, xiao2018spatially, carlini2017towards}. In other words, the goal of this attack is to find an input $x^*$, close to a natural input vector $x$, such that $F(x^*) \neq F(x)$. Despite networks having high levels of accuracy on normal inputs, prior works show that neural networks are extremely susceptible to these well-crafted attacks. Notably, these perturbations also generalize well to different networks as the same perturbations have been shown to be capable of fooling multiple networks~\cite{moosavi2017universal}.% More formally,

In the literature, a large number of works on adversarial examples have been developed for generating stronger methods of producing the adversarial inputs~\cite{akhtar2018threat, tramer2017ensemble}. For instance, the fast gradient sign method (FGSM)~\cite{szegedy2013intriguing} generates the adversarial example in the direction of the sign of the cost function's gradient to produce an adversarial input with very slight perturbation. The jacobian-based saliency maps attack (JSMA) algorithm~\cite{papernot2016limitations} uses the gradients of the learned function, rather than the cost function, to produce a saliency map of the input. The saliency map could indicate whether specific values of the input should be increased or decreased to produce a desired change in the output. Besides, several advanced adversarial attacks have also been proposed recently to compromise specific defense mechanisms~\cite{athalye2018obfuscated} or extend to different network architectures and adversarial scenarios~\cite{papernot2016crafting}.  %\textbf{Need to add one or two sentence summarizing the rest w.}%This algorithm only needed to perturb as little as $4\%$ of the input features to produce the desired misclassification. %with the draw back that the change in the input features is more significant that FSGM making it possible for a human to distinguish the two inputs. %Despite being able to tell the images apart a human observer can still classify the perturbed inputs with high accuracy.

%This paper builds upon these prior works on adversarial samples to extend the adversarial model into the supply chain of neural network devices.

\subsection{Hardware Trojan}

Modern integrated circuit design often involves a number of design houses, fabrications houses, third-party IP, and electronic design automation tools that are all supplied by different vendors. Such horizontal business model makes the security extremely difficult to manage during the supply chain. Any of the parties involved in the process may hold incentives to insert hardware Trojans (i.e., maliciously modify the hardware implementation) into the design. Typically, the hardware Trojan would only be activated by rare trigger conditions such that infected devices can still pass a normal functional test without being detected. Thus, hardware Trojan attack can be a critical threat due to its stealthy nature. A hardware Trojan is usually characterized by the activation mechanism (i.e., trigger) and the effect on the circuit functionality when it is triggered (i.e., payload)~\cite{chakraborty2009hardware}. When the trigger condition is satisfied, the payload will accomplish the objective of the Trojan. In the literature, various types of hardware Trojans have been developed~\cite{chakraborty2009hardware, tehranipoor2010survey, jin2008hardware, xiao2016hardware}. %In addition, techniques for detecting hardware Trojan by using side-channel information suffer from reduced sensitivity toward small Trojans, especially given the relatively large process variations in deep nanometer technologies.

\section{Threat Model and Taxonomy}\label{sec:tax}

In the context of machine learning, the adversary could inject Trojans into the model by maliciously altering its weights so that the neural network will malfunction when the Trojan is triggered. In the literature, several works on neural network software Trojan attacks have been developed~\cite{geigel2013neural, liu2017neural, liu2017trojaning}. From the supply chain perspective, maliciously intended modifications to these devices during the process can further provide attackers with new capabilities of altering the functions of internal neurons and causing adversarial functionality. Hardware Trojans can be inserted into a device during manufacturing by an untrusted semiconductor foundry or through the integration of an untrusted third-party IP. Furthermore, a foundry or even a designer may possibly be pressured by the government to maliciously manipulate the design for overseas products, which can then be weaponized. Therefore, it is of great importance to examine the implication of hardware Trojan on neural networks. In this paper, we expand the attacks on neural networks from the training and application phases to the production phase.% \textbf{Do we have more terminologies to add? If only one, we probably will delete this sentence.}%Since the fields of neural network and hardware Trojan use several same terminologies but representing different meanings, we restrict the use of \emph{activation}, \emph{}

%Neural networks, especially deep neural networks, are usually implicit in the decision process, which makes their security difficult to manage. Besides the adversarial sampling attacks and poisoning attacks, Trojan attack has also been developed to compromise the security of a neural network implemented in software by maliciously altering its weights so that the neural network will malfunction when the Trojan is triggered []. In this paper, we expand the attacks on neural networks from the training and applications phases to the production phase, i.e., the hardware implementation perspective.

\subsection{Threat Model}
Unlike software Trojans, hardware Trojans consider the malicious modification of the original circuitry~\cite{chakraborty2009hardware, tehranipoor2010survey}. An inserted hardware Trojan will change circuit functionality by adding, deleting or modifying the components to wrest control from the original chip owners. As opposed to software Trojans, hardware Trojans would have both capabilities of changing the weights and altering the functionalities of specific neurons depending on the threat model. This difference undoubtedly leads to vastly distinct insertion and design strategies for neural network hardware Trojans. Indeed, given that the hardware Trojan produces new threat models with no equivalent software counterpart, attack and defense scenarios must be first studied.

\begin{figure}[htbp]
\centering
\resizebox{0.5\textwidth}{!}{%
\includegraphics[]{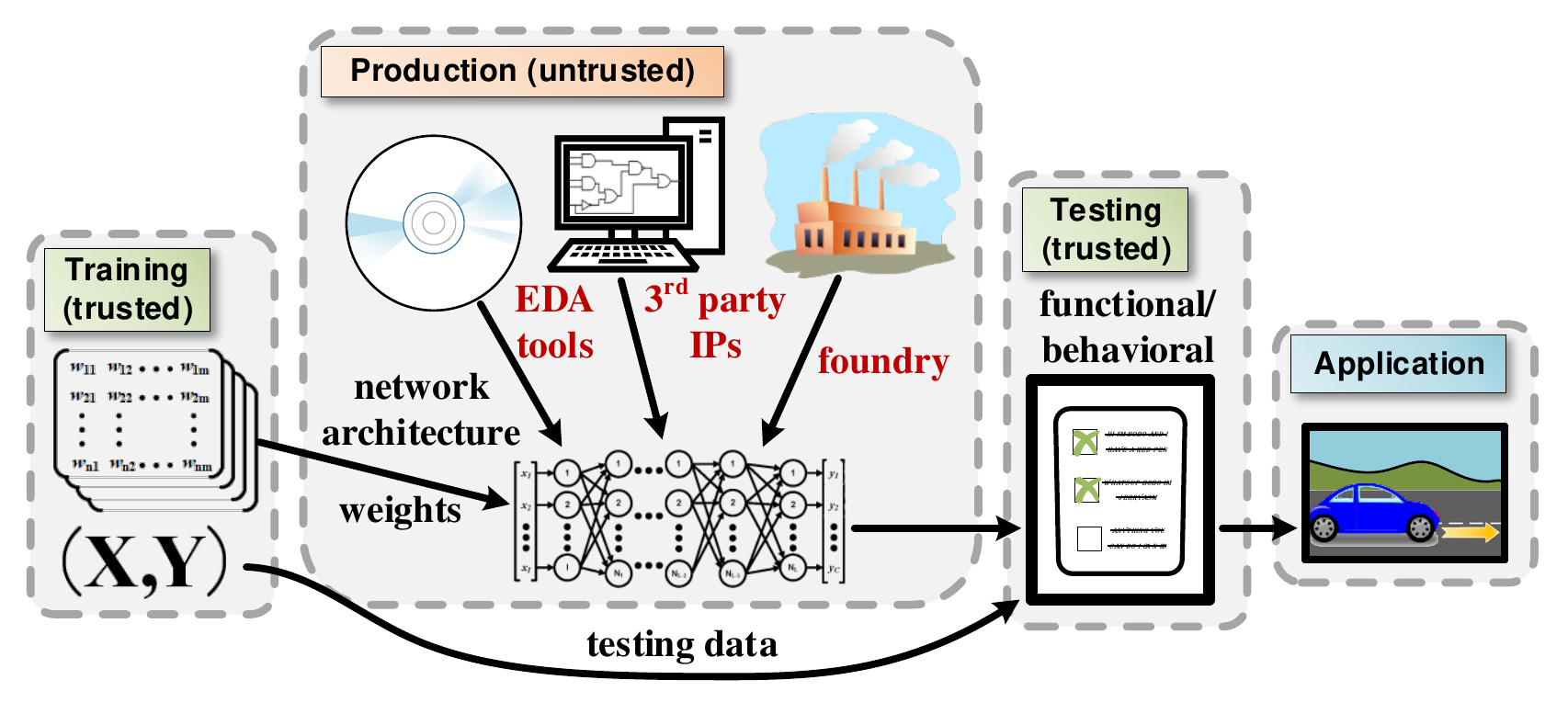}}
\caption{\small The adversarial setting considered in this paper.}
\figlbl{Threatmodel}
\end{figure}

In this paper, we consider a threat model that assumes an adversary is positioned in the supply chain of an integrated circuit containing a well-trained neural network model, as shown in~\figref{Threatmodel}. This threat model is particularly revelent given that many companies wish to use offshore state-of-the-art technologies to remain competitive in the market, especially for neural network devices whose performance are crucial for real-time applications. It is also plausible that, due to potential speed-ups and improvements in power consumption, the designer desires to hard-wire the network parameters. This setting would give the adversary direct knowledge of architecture and all weights associated with the well-trained model. However, the adversary would not have the knowledge of the training or test data.

The objective of the adversary is to insert a hardware Trojan into the original design of the neural network circuit forcing a specific trigger input to be misclassified to either a targeted or an untargeted class. Under this scenario, the adversary can modify both the weights and the functionalities of circuit components prior to shipment. In order to evade detection, the adversary should ensure the hardware Trojan is stealthy enough such that the predictions for the unknown test data are completely unmodified. In addition, the physical footprint of the hardware Trojan must remain sufficiently tiny; thus, the Trojan-injected circuit would be difficult to differentiate from the original "golden circuit". In this paper, we focus on the hardware Trojan attack on neural network circuit components, while we expect the hardware Trojan targeting on the weights would yield a similar impact as the software Trojan or fault injection attack.

\subsection{Expanding the Taxonomy of Neural Network Attacks}
In the literature, taxonomy of attacks on neural networks~\cite{barreno2006can,papernot2016limitations,papernot2016towards} are divided into the four domains: the phase at which the attack is initiated, the goal of the attacker, the scope of the attack, and the attacker's knowledge of the system, as shown in~\figref{Taxonomy}. In particular, the attacks are classified into two phases according to the stages of the neural network: the training phase and the application/inference phase~\cite{papernot2016towards}. An attack in the training phase seeks to take control over the training algorithm by maliciously altering the trained model. On the other hand, an attack in the inference phase attempts to explore possible flaws in the trained model without tampering with the network. Given the threat model we described above, we consider the attacks on the production phase of the well-trained neural network for the first time, to the best of our knowledge. Note that the fault injection attack on the neural network~\cite{liu2017fault} still falls into the inference or application phase. Since the described threat model assumes partial knowledge of the neural network model, the hardware Trojan attack is considered as a greybox attack. In sum, we consider the hardware Trojan attack during the production phase to compromise the reliability of neural networks with both targeted and untargeted scopes in this paper, as circled in~\figref{Taxonomy}. %Besides, the hardware Trojan attack can be either targeted or untargeted according to the objective of the malicious behavior. Finally, the knowledge available to the adversary divides different attacks into whitebox, greybox, or blackbox attacks. .

\begin{figure}[htbp]
\centering
\resizebox{0.5\textwidth}{!}{%
\includegraphics[]{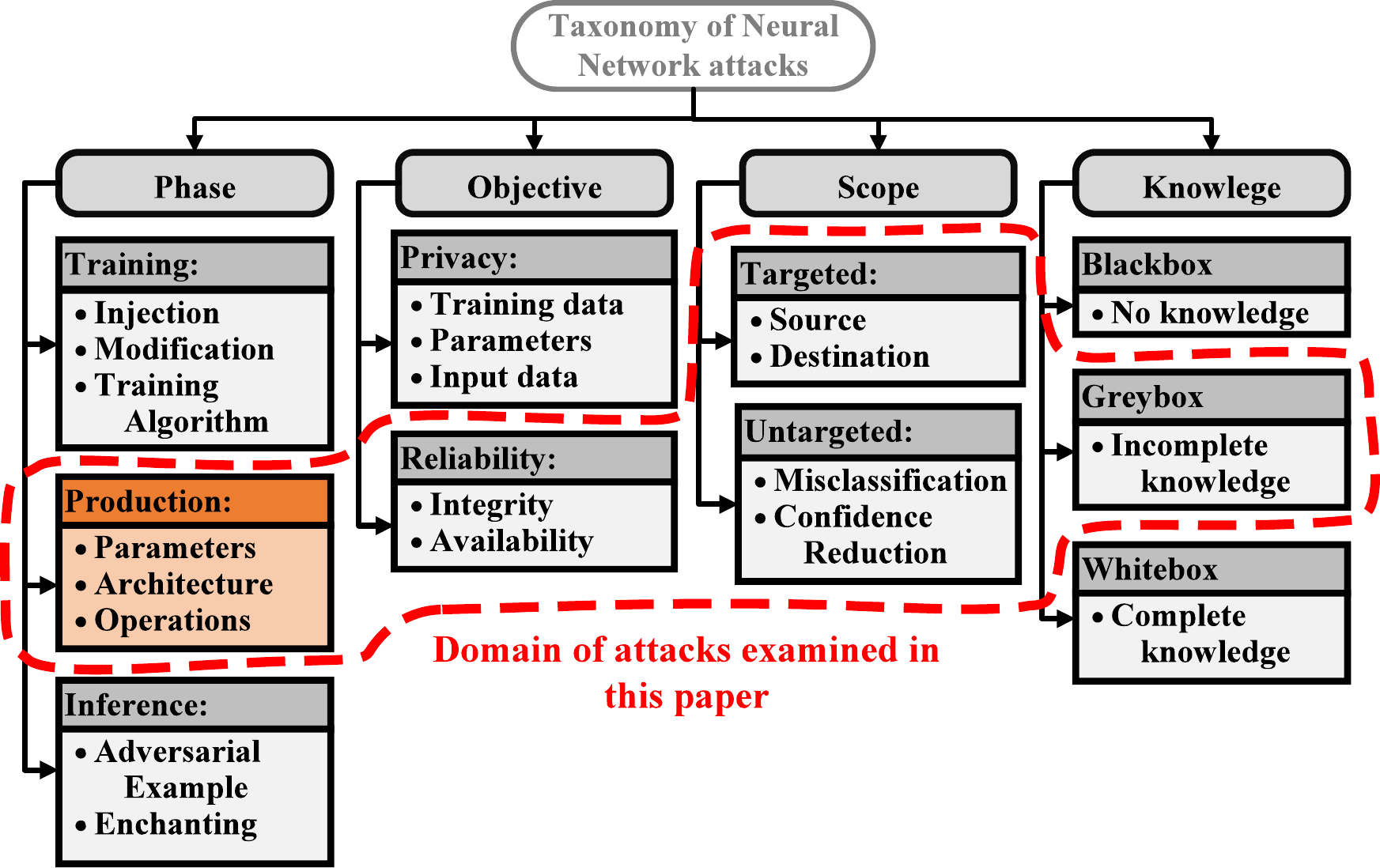}}
\caption{\small The expanded taxonomy of neural network attacks.}
\figlbl{Taxonomy}
\end{figure}

\section{Algorithms for Hardware Trojan Attacks on Neural Networks} \label{sec:over}
%With tradeoffs between stealth and effectiveness dominating the space of hardware Trojans,
In this section, we first build a general framework for inserting hardware Trojans in neural networks under the threat model presented in Section III, and then develop several algorithms for different adversarial scenarios.

\subsection{General Framework}
The proposed framework consists of two main steps: (i) malicious behavior generation, i.e., determining the neuron(s) to inject the Trojan and the corresponding perturbations, and (ii) hardware Trojan implementation, i.e., designing the trigger and payload circuitry. Our proposed methodology provides an adversary the flexibility in selecting the targeted layer of a neural network for injecting hardware Trojan. Without loss of generality, we assume the targeted layer is layer $l$. An example of Trojan-injected neural network is shown in~\figref{NeuralNet}. When the trigger condition is satisfied, the injected neurons will propagate the malicious behaviors to subsequently layers and finally modify the output prediction, as marked in red in~\figref{NeuralNet}. Note that multiple Trojans need to be injected to achieve the attacking objective in most cases, as each operation in the network has a minor effect on the output within the dynamic range $T$, especially for deep neural networks.

\begin{figure}[htb]
\centering
\resizebox{0.45\textwidth}{!}{%
\includegraphics[]{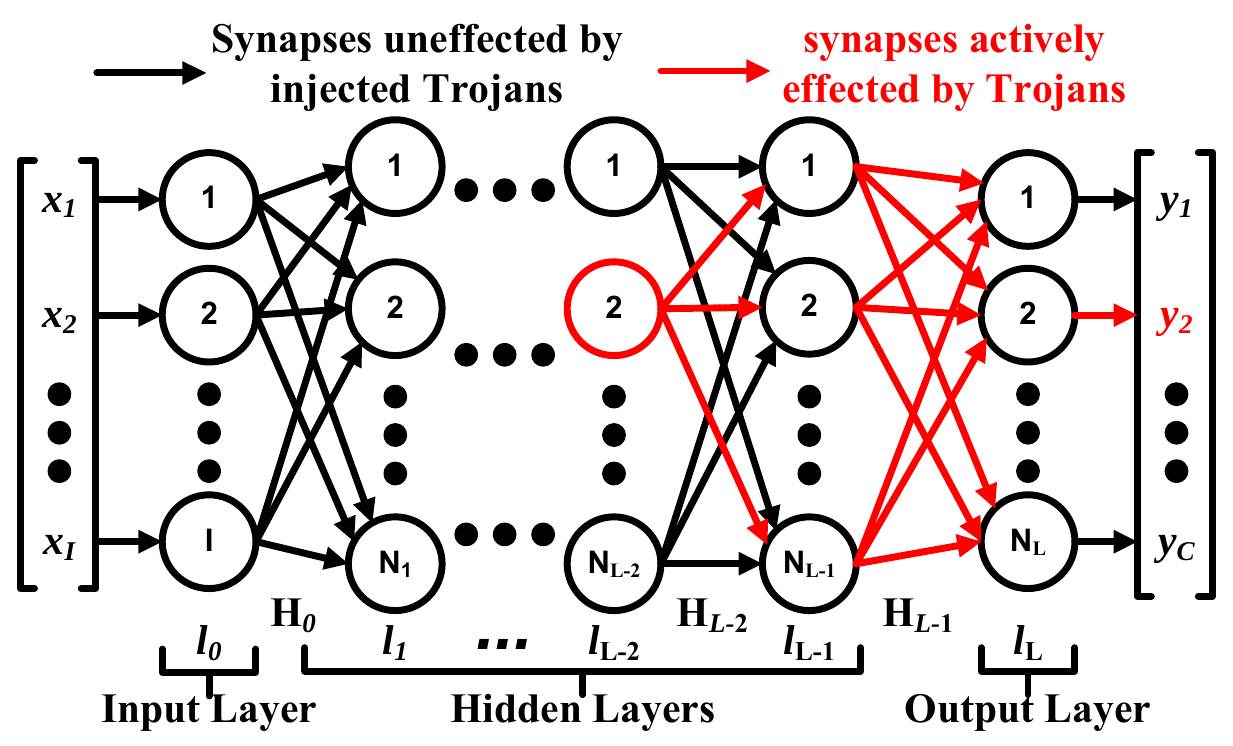}}
\caption{\small A neural network injected with hardware Trojans, the effect of the Trojans is propagating through some neurons but can be filtered out on others.}
\figlbl{NeuralNet}
\end{figure}

%\textbf{not sure what do you want to say here}%In most settings it is highly probably that multiple Trojans will need to be injected, as each operation in the network has a minor effect on the output. To begin the attack, a target layer $l$ should be chosen by the attacker, the insertion of all injected will be on this layer, this both reduces the search space required when inserting trojans and removes the possibility that a cascading effect will occur when a Trojan misfires.

%After a layer is chosen, the adversary divides the network \mathit{?}odel into two-subnetwork models at the chosen layer. Because of the layered structure of neural networks, the equations being implemented have a degree of modularity, allowing us to easily divided any network into sub-network at the layers. Formally, the network model can be represented by the two equations:
Due to the layered structure, a neural network can be divided into sub-networks separated at the layer $l$, which can be expressed as
\begin{equation}
\small F(x) = F_{l+1:L}(H_{l}) \mbox{; } H_{l} = F_{l:l}(H_{l-1}) \mbox{ and } H_{l-1} = F_{0:l-1}(x).
\end{equation}
This modularity is further increased by the natural division of operations inside a network layer. For example, as shown in~\figref{Neuron}, a dense layer is usually composed of multiplication operations followed by an accumulation operation and finally an activation function, plus any additional operations such as pooling. These operations create additional natural break points in which an adversary can inject Trojans.

%To perform the attack, the attacker first needs to select an input trigger, we assume that the adversary already has a input trigger in mind as the any arbitrary choice that can be passed to the network can be used. Then, The network model is decomposed into the two sub-networks, $F_{0:\tilde{l}}(\cdot)$ and $F_{\tilde{l}+1:L}(\cdot)$.

To perform the hardware Trojan attack, the adversary also needs to pick an input trigger $\tilde{x}$. In the proposed framework, the trigger can be chosen arbitrarily or similar to legitimate inputs to achieve a higher degree of stealthiness. Then, we use the input trigger and the functions representing the first two sub-networks to obtain the intermediate values following the first and second subnetworks, i.e., $H_{l-1} = F_{0:l-1}(\tilde{x})$ and $H_{l} = F_{l:l}(H_{l-1})$. We then apply a modified adversarial sampling algorithm with respect to the target layer to find perturbation needed to induce in the layer $l$ to achieve either a targeted or untargeted attack. In order words, the goal is to generate $\tilde{H}_l = H_l + p$ such that $F_{l+1:L}(\tilde{H}_{l})$ is altered as intended, while the perturbation $p$ for each modified neuron is bounded by the dynamic range based on the neural network model. Finally, the Trojan circuitry is designed according to the required perturbations and intermediate value. % \textbf{add this somewhere}%In addition, the norm used in this step would be subject to the , where the goal is to reduce the number of operations to be modified, the l0-norm would be used.

%To perform the attack the network model is decomposed into the three sub-networks, $F_{0:\tau-1}(\cdot)$, $F_{\tau:\tau}(\cdot)$ and $F_{\tau+1:L}(\cdot)$, this essentially isolates the target layer with two sub-networks to either side of it. Once the network model has been decomposed into its sub-networks, the trigger input $ x_{\tau} $ is feed through the first sub-network to generate an intermediate value between the two sub-networks, $ h_{\tau-1}$. This value is feed through the second sub-network, composed of the target layer, to produce the intermediate value, $h_{\tau}$. Once these this second intermediate value is known, A method of adversarial sampling can be modified to indicate both the operations to inject Trojans into and the modification required to make these Trojans effective.

\subsection{Malicious Behavior Generation}
While any approach in the existing literature of producing adversarial examples may be incorporated in the above framework, we choose to develop our approaches based on the JSMA algorithm~\cite{papernot2016limitations}, since it is designed specifically for minimizing the $L0-$norm which could potentially minimize hardware modification for Trojan insertion. The JSMA algorithm generates a Jacobian matrix with respect to the input and then utilizes the Jacobian with a set of rules to build a saliency map. By modifying the rules when constructing the saliency map, different adversarial objectives can be prioritized.% making this adversarial sampling method particularly useful in this setting.

%The modified algorithm for use in this scenario is distinct from the original algorithm as the Jacobian of a sub-network model with respect to $h_{\tau}$. Formally, the modified Jacobian is evaluated as:

As opposed to the original JSMA algorithm, in our proposed method, we modify the Jacobian as presented in Equation \ref{eq:JMat}.
\begin{equation}
\label{eq:JMat}
\small J(H_{l}) = \left[ \dfrac{\delta F_{l+1:L}(H_l)[c]}{\delta H_l} \right]_{c \in C}.
\end{equation}
%The method for computing this matrix is referred to as forward propagating the derivative and is similar to the method of backpropagation used in many training algorithms. The distinction between the two is that 1) calculating the Jacobian is begun at the network output rather than the cost function and 2) the gradients are found with respect to the target layer rather than the network parameters.
Calculating the Jacobian begins at each output and is forward propagated to the target layer using the following modified version of the chain rule.
%\begin{equation}
%\small
%\dfrac{\delta F_{l+1:L}(H_l)[c]}{\delta H_l} = f_L'(W_L \cdot H_{L-1})[c] \cdot W_L \cdot \dfrac{\delta H_{L-1}}{\delta H_l}
%\end{equation}
\begin{equation}
\small \dfrac{\delta H_{l_i}}{\delta H_l} = H_{l_i} \cdot W_{l_i} \cdot \dfrac{\delta H_{{l_i}-1}}{\delta H_l}
\end{equation}
Each column in the Jacobian corresponds to a specific output while each row is linked to a specific neuron in the targeted layer. This is distinct from the original algorithm as the rows of the original Jacobian matrix were linked to the input image. The element at the intersections of these rows and columns of this matrix indicate the strength of the correlation between the nueron/output pair. In this way, each entry of the Jacobian matrix indicates the correlation under the $L0-$norm between the output classification and the intermediate value. It should be noted that each neuron is often tied to multiple outputs with varying strengths and so selecting the neurons should be done through a strict set of rules.

Consequently, a saliency map can be generated using the Jacobian matrix based on the specific goal of the adversary. An attacker with a targeted scope seeks to accomplish the goal in a specific way, while the untargeted scope simply attempts to cause the specified input to misclassify to any other classes. In our methodology, we modify the rules for building the saliency map to incorporate both scopes. For instance, in an untargeted attack, the difference between the negative values in the column corresponding to the predicted class, $ dx^-_p $, and the positive values from every other column, $ dx^+_{i\neq p}$, can be used to form the saliency map:
\begin{equation} \label{eq:SMapUntarg}
\small S(x)[i]=\beta_p \left|dx^-_p\right| + \beta_s \sum_{i \neq p}{{dx^+_i}}.
\end{equation}
Each entry in this map essentially indicates  the effectiveness of simultaneously achieving the primary goal (i.e. decreasing the confidence of the predicted class) and the secondary goal (increasing the probability of a different class) by modifying the corresponding neuron. To gain optimal results in specific adversarial settings, $\beta_p$ and $\beta_s$ are introduced to weight the primary and secondary goals.

However, the goal in a targeted attack is to decrease the probability of the targeted class over that of the predicted class. Consequently, decreasing the probability of the currently predicted class remains the primary goal but the attack also incorporates an auxiliary goal of increasing the confidence of the target class. We also imposing a secondary goal of keeping the remaining probabilities low. Thus, we formulate a saliency map using Equation (\ref{eq:SMapTarg}).
\begin{equation} \label{eq:SMapTarg}
\small S(x)[i]= \beta_p \left|dx^-_p\right| + \beta_a dx^+_t + \beta_s \sum_{i \neq t,p}{\left|{dx^-_i}\right|},
\end{equation}
Here we include three constants; $\beta_p$, $\beta_a$ and $\beta_s$, to weight the primary, auxiliary and secondary goals to gain optimal results in specific adversarial settings.

%In this setting keeping the confidences of the other untargeted classes is not as large a priority as the others and so w can be used to decrease the dependence on this constraint.

To find the modification needed in ${H_l}$, we perturb the operation associated with the largest values in the vector and modify them according to the adversarial objective. The magnitude of the perturbation should be bounded by the dynamic range $T$ in the original neural network. For example, the value after the Trojan-injected neuron should be bounded between -1 and 1 if the activation is $tanh$. A $ReLU$ activation function leads to a theoretically unbounded upper limit; however, in a practical real world attack any modifications would be limited due to the physical representation of the values. For the bounded attack, we use a bounding list, $L$, to lock neurons that cannot be further altered in the desired direction. We present the algorithms for the untargeted attack and the targeted attack in Algorithm 1 and Algorithm 2, respectively.  % In many settings, as when modifying . In this case, bounding the modifications between these bounds would limit the possibility of abnormal behavior being observed in the circuit. Below illiterate two different algorithms for implementing these attacks with different goals and constraints in.

%$H_{l}$ and the third sub-network are used to generate an adversarial perturbation $p$ that forces misclassifications at the target layer. $H_{l-1}$ is used to generate the Trojan trigger circuitry and $p$ for the Trojan payload. In the case of bounding a list is used to lock neurons which should not be modified.

\begin{algorithm}[H]
\label{frame}
\caption{\textbf{Untargeted Hardware Trojan Attack}}
\label{Alg1}
\small
\begin{algorithmic}[1]
    \REQUIRE $ F(\cdot), \tilde{x}, T ,l$
    \STATE $F(\cdot) \rightarrow F_{0:{l-1}}(\cdot) \mbox{, } F_{l:{l}}(\cdot)  \mbox{ and }  F_{l+1:L}(\cdot) $
    \STATE $H_{l-1} = F_{0:{l-1}}(\tilde{x}) $
    \STATE $\tilde{H}_l = H_{l} = F_{l:l}(H_{l-1}) $
    %\STATE $\tilde{H}_l =H_l$
    \STATE $y_p = F_{l+1:L}(H_{l}) $
    \STATE $L = [$ $]$
    %feed h_l part way through?%
    \WHILE{$ F_{{l}+1:L}(\tilde{H}_l) = y_p \mbox{ and } \left|| p \right|| < T $}
        \STATE forward propagate $J(\tilde{H}_l)$
        \STATE $S = Untargeted\_SM(J(\tilde{H}_l),y_p),$ using Equation (\ref{eq:SMapUntarg})
        \STATE increase $\tilde{h}_n = argmax(S)$
        \STATE $p = \tilde{H}_l-H_l$
        \IF {$ \left| \tilde{h}_n \right| $ exceeds  $ T $}
            \STATE $L.append(h_{n})$
        \ENDIF
    \ENDWHILE
    \STATE generate trigger design based on $H_{l-1}$
    \STATE generate payload design using $p$
\end{algorithmic}
\end{algorithm}
\vspace{-1em}

\begin{algorithm}[H]
\label{frame}
\caption{\textbf{Targeted Hardware Trojan Attack}}
\label{Alg1}
\small
\begin{algorithmic}[1]
    \REQUIRE $ F(\cdot), \tilde{x}, \tilde{y}, T, l$
    \STATE $F(\cdot) \rightarrow F_{0:{l-1}}(\cdot) \mbox{, } F_{l:{l}}(\cdot)  \mbox{ and }  F_{l+1:L}(\cdot) $
    \STATE $H_{l-1} = F_{0:{l-1}}(\tilde{x}) $
    \STATE $\tilde{H}_l =H_{l} = F_{l:l}(H_{l-1}) $
    %\STATE $\tilde{H}_l =H_l$
    \STATE $L = [$ $]$
    %feed h_l part way through?%
    \WHILE{$ F_{{l}+1:L}(\tilde{H}_l) \neq \tilde{y} \mbox{ and } \left|| p \right|| < T $}
        \STATE forward propagate $J(\tilde{H}_l)$
        \STATE $S = Targeted\_SM(J(\tilde{H}_l),\tilde{y}),$ using Equation (\ref{eq:SMapTarg})
        \STATE increase $\tilde{h}_n = argmax(S)$
        \STATE $p = \tilde{H}_l-H_l$
        \IF {$ \left| \tilde{h}_n \right| $ exceeds  $ T $}
            \STATE $L.append(h_{n})$
        \ENDIF
    \ENDWHILE
    \STATE generate Trojan trigger design based on $H_{l-1}$
    \STATE generate Trojan payload design using $p$
\end{algorithmic}
\end{algorithm}
\vspace{-1em}

In addition to the original saliency map that indicates which neuron outputs to increase, we implemented the targeted attack with a second saliency map that indicates which neuron outputs to decrease. This slight modification allowed our implementation to mount attacks more quickly and efficiently than when utilizing only the single saliency map above.

\subsection{Hardware Trojan Implementation}
The implementation of the hardware Trojan design is highly dependent on the specific neural network architecture and the injected component of choice. Here, we only lay the groundwork and describe several possible designs. Note that other hardware Trojan designs of different types but with similar functionalities can also be incorporated into the proposed framework.

%\subsubsection{Trigger Design}
The trigger of the hardware Trojan should be designed based on the internal state of the injected location, i.e., the produced ${H}_{l-1}$ when feeding $\tilde{x}$ through $F_{0:{l-1}}(\cdot)$. In addition, the triggerability must be extremely low to ensure the stealthiness of the hardware Trojan. A combinational circuit can be used to trigger the Trojan when even $H_{l-1}$ closely resembles $\tilde{H}_{l-1}$. In the proposed framework, the payload should be designed to achieve the needed perturbation $p(H_{l})$ obtained from the malicious behavior generation step. We can either use a multiplexer logic which selects output of malicious logic only when the Trojan is activated, or alter the internal structure of the certain operations to inject malicious behavior. For instance, several multipliers can be modified to produce rare outputs given the vector of $H_{l-1}$. We can also target on the activation function of each layer to directly alter $H_l$ after the layer. Although our algorithms for malicious behavior generation are designed to minimize the hardware modification, we must still be careful in selecting the payload design such that the magnitude of change (e.g., the difference in side-channel leakage) is small enough to evade existing hardware Trojan detection techniques. The simplified block diagrams of two possible hardware Trojan designs are shown in~\figref{Trojan}. %Beyond this many other possible Payload

\begin{figure}[htb]
\centering
\resizebox{0.4\textwidth}{!}{%
\includegraphics[]{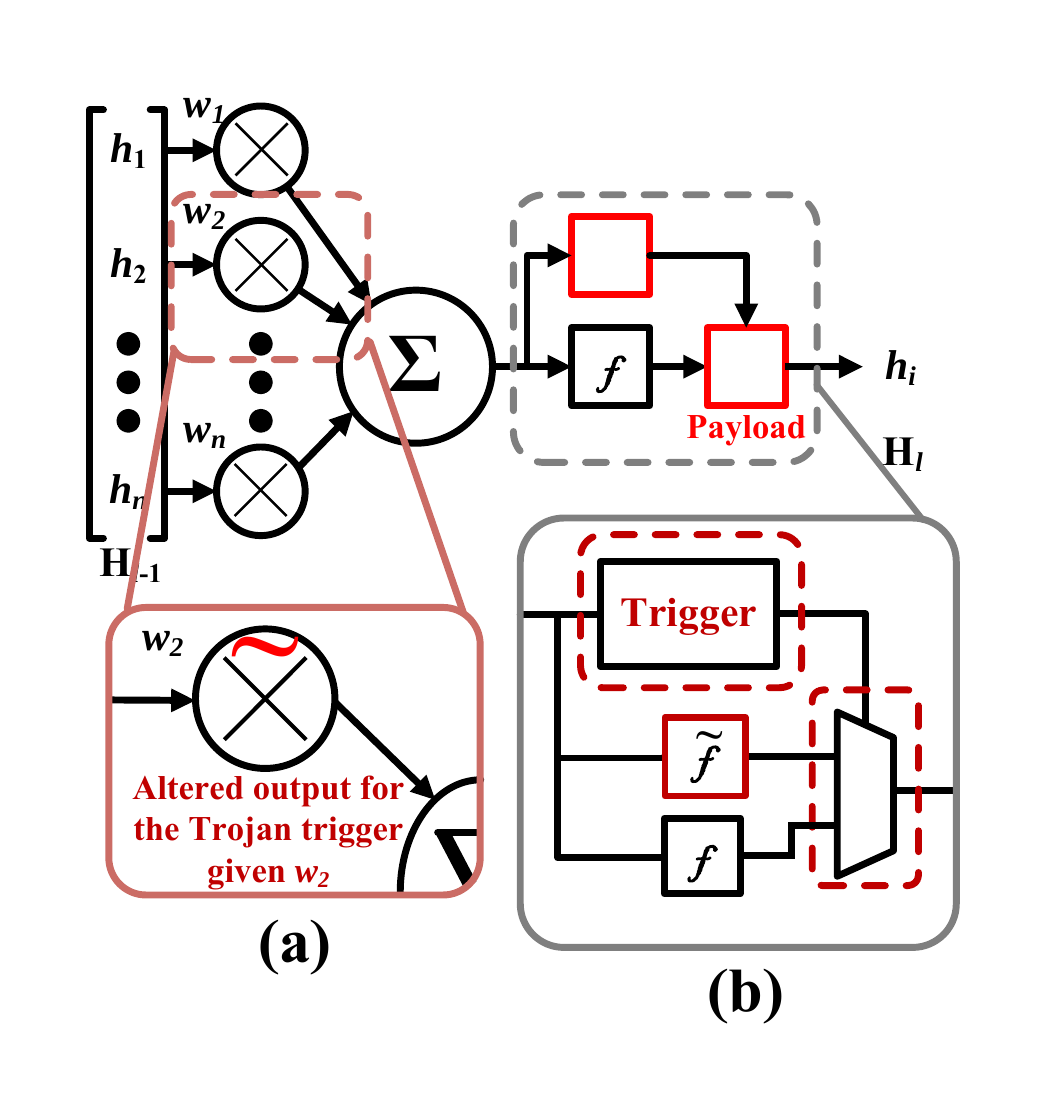}}
\caption{\small Simplified representations of two possible hardware Trojan designs on a neural network.}
\figlbl{Trojan}
\vspace{-1em}
\end{figure}

\section{Experimental Results}\label{sec:exp}
\subsection{Datasets and Neural Network Models}
We use MNIST and CIFAR10 datasets to evaluate the proposed methodology. Both datasets are composed of 10 mutually exclusive classes. The detailed hyperparameters of models that we implemented in our experiments are summarized in Table \ref{table:networks}. We pre-train the networks to achieve test accuracies above $99\%$ and $80\%$ on the MNIST and the CIFAR10 datasets, respectively, which are consistent with the state-of-the-art. We then consider these models as the original benign models to inject hardware Trojan using the proposed algorithms. We run the experiments on a cluster node with NVIDIA Tesla GPUs. The training of the neural networks is implemented using Tensorflow~\cite{tensorflow}.

\begin{table}[hbt]
\centering
\begin{small}
   \caption{Summaries of network architectures}
   \label{table:networks}
   \begin{tabular}{|c || c| c| c|  c|}
      \hline
       & \multicolumn{2}{c|}{MNIST} & \multicolumn{2}{c|}{CIFAR-10}\\
      \hline
       layer& type & \# neurons & type & \# neurons \\
       \hline
       \hline
       1& conv 20& 15680 & conv 32 & 28800 \\
       \hline
       2& conv/max 40 & 31360 & conv/max 64 & 50176 \\
       \hline
       3& conv 60& 11760 & conv/max 128 & 18432 \\
       \hline
       4& conv/max 80 & 15680 & conv/max 128& 2048 \\
       \hline
       5& conv 120 & 5880 & dense & 1024 \\
       \hline
       6& dense & 150 & dense & 180 \\
       \hline
       7& dense & 10 & dense & 10 \\
       \hline
       \multicolumn{5}{l}{\footnotesize max-pooling size: 2x2, kernel size: 3x3}

   \end{tabular}
   \end{small}
\end{table}

\subsection{Adversarial Scenarios}

\subsubsection{Scope of the Attack}
We evaluate both \textbf{targeted} and \textbf{untargeted} hardware Trojan attacks on the above neural networks. In our experiments, we use every other class of each dataset excepted the correct label as the targeted class for the targeted attack. While for the untargeted attack, we simply attempt to alter the prediction without a targeted class.

\subsubsection{Input Trigger Selection}
In our experiments, we consider two different input trigger designs, i.e., \textbf{well-crafted} and \textbf{random} input triggers. Well-crafted input triggers are intended to achieve higher degrees of stealthiness against human observers by making the trigger very close to legitimate inputs. In order to ensure the similarity of the well-crafted input trigger, $\tilde{x}$, to the test images, we randomly pull a single instance from the test set and form a new set for testing with the remaining samples. Randomized images adhering to the standards of the datasets are used as random input trigger.

\subsubsection{Payload Constraint}
The possible magnitude of perturbation generated by the payload is constrained by the dynamic range of the original benign model. We use \emph{ReLU} as the activation function on each layer when illustrating the \textbf{unbounded} scenario, while using \emph{tanh} as the activation function for the \textbf{bounded} scenario.

\subsubsection{Targeted Layer}

We examine the performance of hardware Trojan attacks on all hidden layers and the output layer. We show that the proposed method could inject hardware Trojans into any layer to generate malicious behavior and compare the effectiveness and stealthiness of different layers of choice.

\begin{table*}[hbt]
\centering
\begin{small}
   \caption{Random input triggers for targeted attacks}
   \label{table:random}
   \begin{tabular}{|c | c| c| c| c| c| c| c| c|}
      \hline
       & \multicolumn{4}{c|}{MNIST} & \multicolumn{4}{c|}{CIFAR-10} \\
      \hline
       & \multicolumn{2}{c|}{bound} & \multicolumn{2}{c|}{unbound} &\multicolumn{2}{c|}{bound} & \multicolumn{2}{c|}{unbound}\\
      \hline
      layer & $mfn$ (\%)& $eff$ (\%)& $mfn$ (\%)& $eff$ (\%)& $mfn$ (\%)& $eff$ (\%)& $mfn$ (\%)& $eff$ (\%)\\
      \hline
      1& 0.19 & 100 & 0.06 & 100  & 0.21 & 100 & 0.09 & 100 \\
      \hline
      2 & 0.10 & 98 & 0.04 & 100 & 0.14 & 100 & 0.06 & 100 \\
      \hline
      3 & 0.39 & 95 & 0.14 & 100 & 0.15 & 100 & 0.09 & 100 \\
      \hline
      4 & 0.22 & 100 & 0.07 & 100 & 0.81 & 100 & 0.54 & 100 \\
      \hline
      5 & 1.51 & 96 & 0.09 & 98 & 4.57 & 100 & 0.34 & 100 \\
      \hline
      6 & 8.13 & 100 & 2.71 & 100 & 13.53 & 100 & 1.69 & 100 \\
      \hline
      7 & 21.20 & 100 & 30.53 & 100  & 20.20 & 100 & 21.80 & 100 \\
      \hline

   \end{tabular}
\end{small}
\end{table*}

\subsection{Metrics}
The successfulness of a hardware Trojan attack is measured by its capabilities of altering the predictions and evading detection. In this work, we use the effectiveness and the stealthiness to evaluate these metrics. Additionally, we use the number of modified neurons to show the magnitude of change from the hardware implementation perspective. In sum, we define the following metrics:
\begin{itemize}
  \item \textbf{Effectiveness ($eff$)} is the percentage of input triggers classified as the intended label, in either targeted or untargeted scenarios. An effective hardware Trojan attack should yield a high value (i.e., close to $100\%$) of this metric.
  \item \textbf{Stealthiness ($stl$)} is the percentage of the outputs obtained from the Trojan-injected network matches the predictions of the benign network. Ideally, this metric should be $100\%$ to avoid detection by test data that are unknown to the adversary, when returned to the designer.
  \item Number of \textbf{modified neurons ($mfn$)} is the average number of neurons that are modified to generate the desired malicious behavior. This metric directly correlates to the amount of hardware modifications needed to implement the attack. This metric is also reported as the percentage of neuron modified in the targeted layer needed to achieve a misclassification.
\end{itemize}

\subsection{Results and Discussion}

\begin{figure}[htbp]
\centering
\resizebox{0.3\textwidth}{!}{%
\includegraphics[]{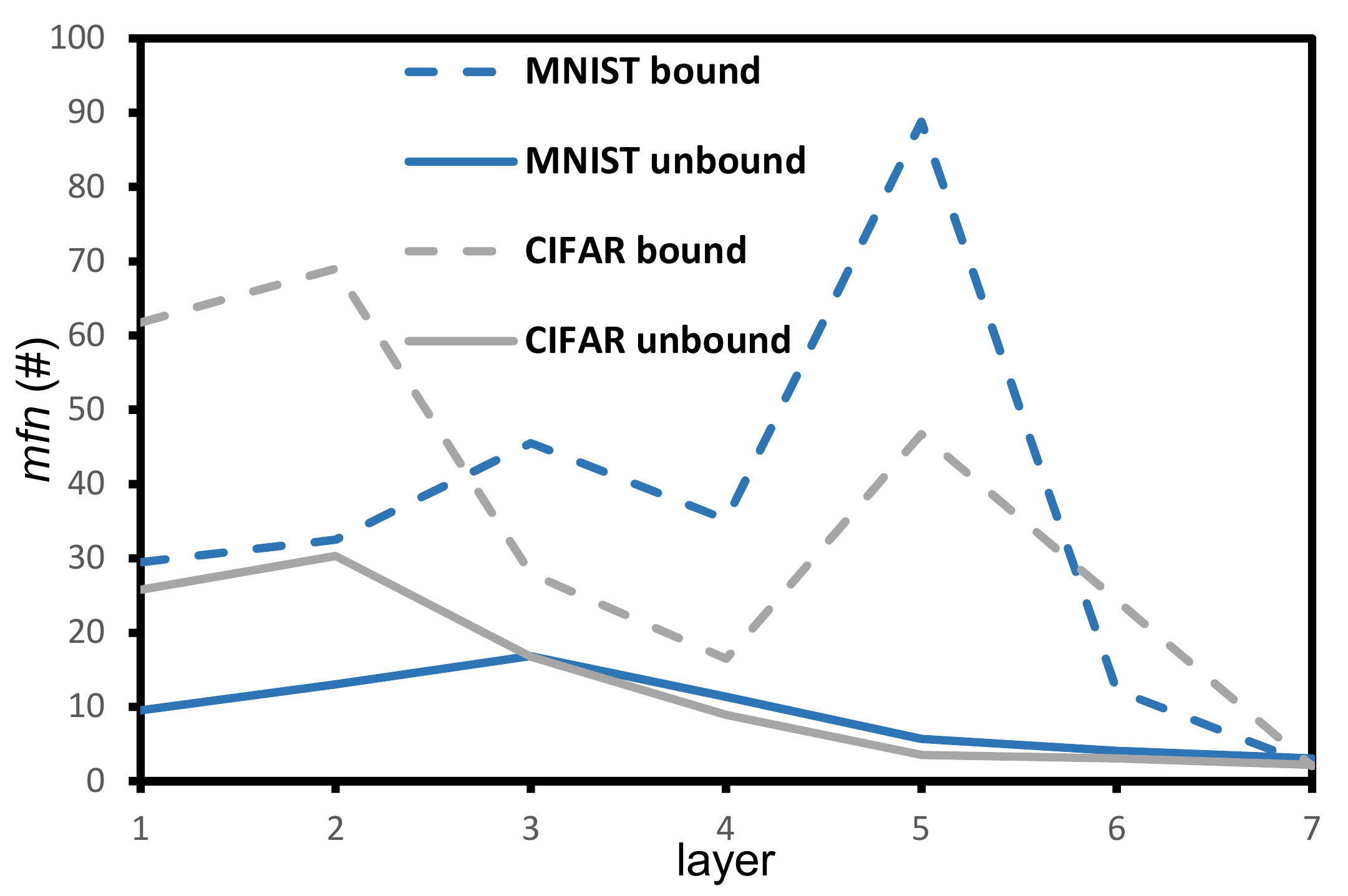}}
\caption{\small Number of modified neurons per layer given random trigger inputs in the targeted adversarial setting.}
\label{BUB}
\vspace{-1em}
\end{figure}

\begin{figure}[htb]
\centering
\resizebox{0.3\textwidth}{!}{%
\includegraphics[]{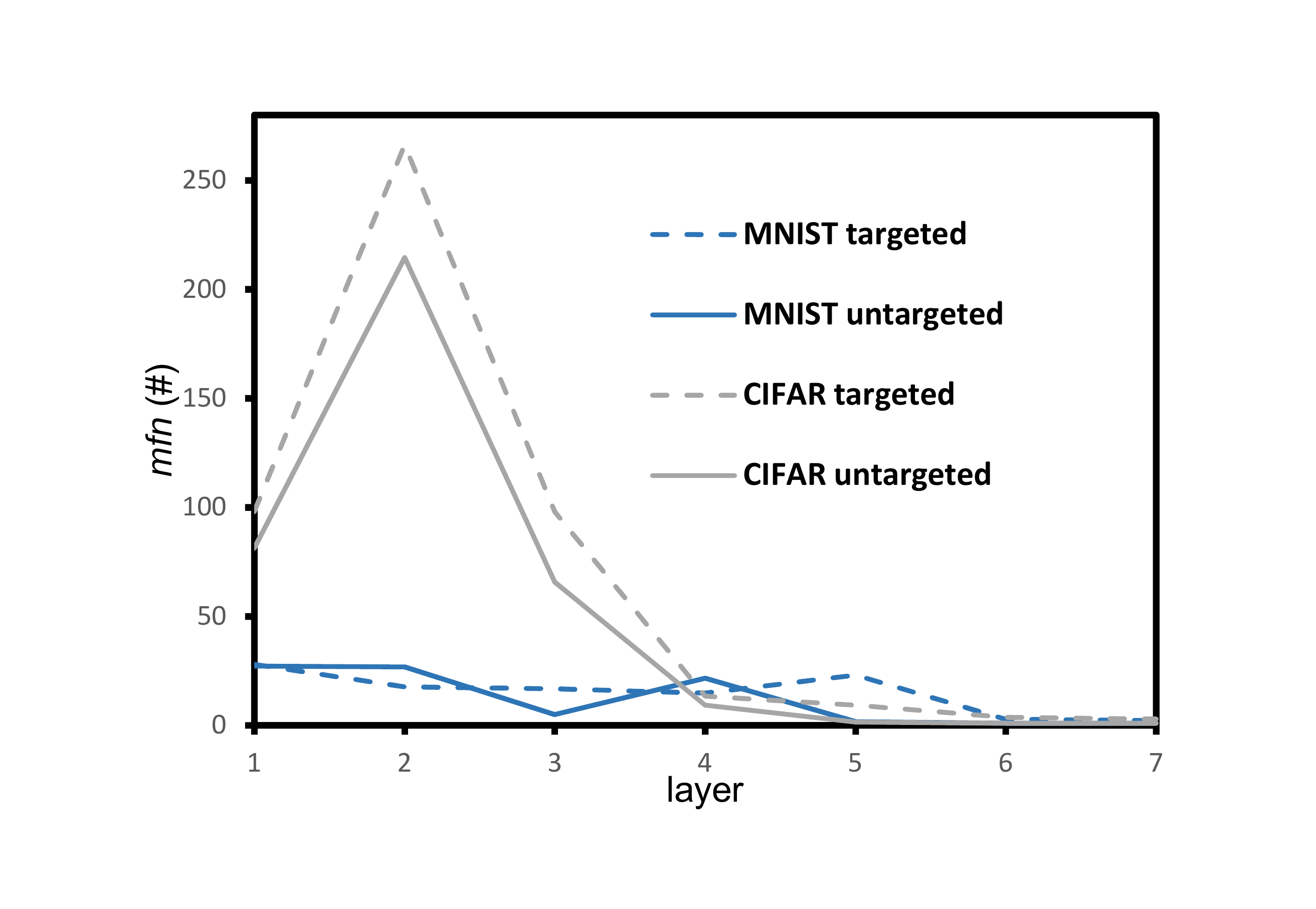}}
\caption{\small Number of modified neurons per layer given well-crafted trigger inputs in the unbounded adversarial setting. }
\label{TUT}
%\vspace{-1em}
\end{figure}

%Intro
The results of our experiments are summarized in Table \ref{table:random} and Table \ref{table:similar}. Note that each experiment is conducted at least 1000 iterations and the averages are presented. In addition, we test the stealthiness of the proposed methods using the test data of each dataset. \textbf{Our experimental results show that the proposed algorithms achieve $100\%$ stealthiness for both datasets under all adversarial scenarios.} It can be seen from both Tables \ref{table:random} and \ref{table:similar} that the percentage of modified neurons increases towards the latter layers of both networks. However, if we compare the absolute value of modified neurons, as shown in Figures \ref{BUB} and \ref{TUT}, latter layers actually require significantly less neurons to be modified to inject malicious behavior. Thus, injecting into neurons in latter layers could result in higher impact to the output, which conforms to our expectation. Note that the lowest possible percentage of modified neurons is $10\%$ for the output layer, since it has a total of 10 neurons in both networks.

%the final layers
 %The untargeted attacks reach this lower bound as they are prioritized to decrease the predicted class while the targeted attacks do not as they equally prioritize increasing the confidence of a targeted class and decreasing the probability of the predicted class.

%bound vs. unbound
We first use random input triggers to mount the hardware Trojan attacks under the targeted adversarial scenarios. The results are presented in Table \ref{table:random} and Fig. \ref{BUB}. It can be seen that the targeted attack under the unbounded scenario is stronger than the bounded scenario, as it requires less neurons to be modified. In other words, different neural network designs also lead to different levels of security from the hardware perspective. It also appears that both of these attacks perform well reaching near $100\%$ effectiveness on all layers while modifying only a small sample of the neurons. For example, our method could effectively classify a random input trigger as a specified class on the MNIST dataset by injecting hardware Trojans into $0.04\%$, on average, of neurons in the 2nd hidden layer the neural network.

\begin{table*}[hbt]
\centering
\begin{small}
   \caption{Well-crafted input triggers under the unbounded scenario}
   \label{table:similar}
   \begin{tabular}{|c | c| c| c| c| c| c| c| c|}
      \hline
       & \multicolumn{4}{c|}{MNIST} & \multicolumn{4}{c|}{CIFAR-10} \\
      \hline
       & \multicolumn{2}{c|}{targeted} & \multicolumn{2}{c|}{untargeted} &\multicolumn{2}{c|}{targeted} & \multicolumn{2}{c|}{untargeted}\\
      \hline
      layer & $mfn$ (\%)& $eff$ (\%)& $mfn$ (\%)& $eff$ (\%)& $mfn$ (\%)& $eff$ (\%)& $mfn$ (\%)& $eff$ (\%)\\
      \hline
      1& 0.18 & 100 & 0.17 & 100  & 0.34 & 97 & 0.28 & 100 \\
      \hline
      2 & 0.06 & 98 & 0.08 & 100 & 0.53 & 100 & 0.43 & 100 \\
      \hline
      3 & 0.14 & 98 & 0.04 & 100 & 0.53 & 100 & 0.36 & 100 \\
      \hline
      4 & 0.09 & 100 & 0.13 & 100 & 0.66 & 100 & 0.45 & 100 \\
      \hline
      5 & 0.39 & 99 & 0.03 & 100 & 0.90 & 100 & 0.15 & 100 \\
      \hline
      6 & 1.83 & 100 & 0.67 & 100 & 2.11 & 97 & 0.55 & 100 \\
      \hline
      7 & 22.17 & 100 & 10.00 & 100  & 29.30 & 97 & 10.00 & 100 \\
      \hline

   \end{tabular}
   \end{small}
\end{table*}

%targeted vs. untargeted
We next evaluate the performance of well-crafted input triggers on the datasets under the unbounded scenario. The results are illustrated in Table \ref{table:similar} and Fig. \ref{TUT}. It can be seen that these attacks also achieve very high effectiveness, while modifying a small percentage of the neurons. For instance, our algorithm can effectively alter the classification of a well-crafted input image in an untargeted scenario while only altering, on average, $0.03\%$ of the neurons in the 5th layer of an MNIST classifier. Under this scenario, it can be observed that the untargeted attack usually requires less modifications than targeted attack, since it has the flexibility to select the easiest malicious output.% Despite the two scenarios appearing to give a comparable $mfn$ in the earlier layers of the MNIST network, .

% per and tot mfn
When comparing the results between the MNIST and CIFAR10 classifiers under the same adversarial settings, we can observe that attacks on the CIFAR10 classifier in general require larger percentage of neurons to be modified. This is further compounded by the fact that the majority of the layers in the CIFAR10 classifier have more neurons than the corresponding MNIST classifier. For example, when targeting on the 2nd layer, our algorithm only needs to modify less than $50$ neurons of the MNIST classifier, while over $200$ neurons have to be altered in the CIFAR10 classifier. %This difference is especially pronounced in the targeted and untargeted attacks on the early layers of the two classifiers. To illustrate, these attacks on the 2nd layer of the MNIST classifier both need an absolute $mfn$ of less than $50$ while requiring over $200$ in the CIFAR10 classifier.

%difference between layer types
%When attempting to compare the results of these experiments over both the MNIST and CIFAR10 classifiers it is important to note that the two networks were generated with different architectures, seen in Table \ref{table:networks}. These differences in architecture increase the difficulty in comparing results between the two networks. For example, the 5th layer of the MNIST classifier is a convolutional layer with 5 times more neurons than the densely connected layer found in the CIFAR10 classifier. Observing Fig. \ref{BUB} we can see that the bounded attacks on this layer in the CIFAR10 classifier have a significantly higher absolute $mfn$ than the unbounded attack on the same layer. Despite the MNIST classifier typically requiring a smaller absolute $mfn$, this trend is even more pronounced in the bound and unbound attacks on its 5th layer. This pattern can be seen to continue, to a smaller degree, in the targeted and untargeted attacks on the same layer in Fig. \ref{TUT}. It is likely that if applied to more complex architectures such as inception networks these differences could become even more pronounced.

%compare random and well-crafted
Finally, we observe that the adversary's choice of input trigger affects the strength of the attack. By comparing experimental results between the well-crafted and random input triggers of the CIFAR10 classifier, it is apparent that the attacks based on well-crafted input triggers require more modifications. Specifically, attacks on the second layer of the network require almost 9 times more modifications with well-crafted input triggers, compared to random input triggers.  Thus, random input trigger could achieve higher stealthiness.

%I have included another graph in the figs folder to illustrate this point but I thought including it in the paper redundant
%Finally, we observe that the adversary's choice of input trigger effects the strength of the attack. As both well-crafted and random input trigger types were utilized in unbound targeted attacks, we compare the two trigger types under this adversarial setting. The most obvious differences between the input triggers can be seen by comparing results of the attack in the CIFAR10 classifier. It is apparent that the attacks generated from the well-crafted input triggers require more modifications compared to attacks on random input triggers. Specifically, layer two of this network requires almost 9 times more modifications when generating attacks with well-crafted input triggers. While less obvious it can be seen that the MNIST classifier also follows the same trend.

\section{Potential Defense Techniques} \label{sec:def}
In this section, we briefly discuss possible defense techniques against the hardware Trojan attack on neural networks. This type of attack using the proposed methodology will inject malicious behavior with an extremely low trigger rate into the original benign model by modifying the hardware implementation. Although normal test data are very unlikely to discover the malicious behavior, defense strategies from both the hardware and the neural network algorithmic perspectives can still be potentially utilized to improve the resilience of a neural network model against such attacks.

On the one hand, various hardware Trojan detection methods have been developed in the literature, including but not limited to optical inspection, logical testing, side-channel analysis, and run-time monitoring~\cite{narasimhan2012hardware, narasimhan2013hardware}. Most of these techniques require a "golden circuit" and rely on a relatively significant difference between the "golden circuit" and the Trojan-injected circuit. However, these techniques such as detection using side-channel information suffer from reduced sensitivity toward small Trojans, especially given the relatively large process variations in deep nanometer technologies~\cite{narasimhan2013hardware}. Since our algorithms attempt to minimize the hardware modification, which has also been verified by our experimental results, we expect such hardware Trojan detection methods would be ineffective for defeating the proposed attack. In addition, run-time monitoring techniques are usually very expensive or incurring significant resource overheads~\cite{bhunia2013protection}. Preventative methods have also been proposed to make hardware Trojan injection more difficult and non-functional, such as hardware obfuscation~\cite{chakraborty2009security} and split manufacturing~\cite{xiao2015efficient}. However, given the modularized operations of neural networks, the degree of ambiguity these methods could create is extremely limited.

%Detection methods for hardware trojans can be divided into three categories: destructive, invasive and non-invasive. As the name implies destructive methods decompose an IC in an attempt to find hidden trojans. These methods are largely disregarded as they have a large economic cost and can only ensure that no trojan is hidden in the ICs that are destroyed and not those which were not destroyed. Invasive detection methods involve adding circuitry to an IC to aid in the detecting of trojans. For example adding testing points to internal nodes critical to the ICs performance can give an assurance that these nodes atleast are un modified. Non-invasive methods attempt detect Trojans through the use of either logical or side-channel tests. Logical tests on a circuit feed inputs though the IC and observe the logical response, while side-channel tests do the same thing but observe other key parameters of the circuit, power consumption and timing, if any abnormalities out side the bounds of those produced by normal noise strongly hint that a circuit could be infected.

On the other hand, although no prior work has studied hardware Trojans on neural networks, defense strategies against adversarial examples might possibly be extended to improve the robustness of neural network models against hardware Trojan injection. Recently, various methods have been proposed to mitigate the effects of adversarial examples by modifying the training algorithm or the network, or using external add-ons. For example, adversarial training manually inserts correctly labeled adversarial samples into the original training data to improve the robustness of the model~\cite{goodfellow2014explaining}. Besides, generative adversarial net (GAN) based approaches utilize external discriminative network to improve the security by classifying both original training samples and adversarial samples generated by the generative network into the correct classes~\cite{samangouei2018defense}. However, the adversarial samples will be much harder to control when applying these methods against hardware Trojan attacks, as the Trojans are injected into hidden layers as opposed to manipulating the input samples. Some other more advanced yet complex techniques, such as the ensemble methods that generate multiple versions of a classifier with differing network architectures~\cite{abbasi2017robustness, tramer2017ensemble}, are cost-prohibitive to implement on hardware.  %While this method achieves strong protection against simple attack algorithms such as FGSM, it can be easily compromised by more advanced blackbox attacks.

In our opinion, we believe one feasible protection method entails the combination of both adversarial training and hardware Trojan detection. Before production, adversarial training can be used to improve the robustness, which could potentially lead to a significant increase in the number of neurons need to be modified to inject the intended malicious behavior. Consequently, the magnitude of change when the injected hardware Trojan is active might grow sufficiently large to be discovered by hardware Trojan detection methods such as side-channel based detection. Our ongoing work includes the investigation of this combined defense strategy with various detection approaches against the proposed hardware Trojan attack framework.

%Possible defenses against attacks of this nature on neural networks could come from two angles, those which attempt to defend the physical hardware and those which defend against adversarial example attacks. As Adversarial examples are a fairly recent discovery the topic is in constant flux with new defenses and new attacks which counter these defenses constantly appearing. Some techniques such as Ensemble Adversarial Training and bolstering defenses with Generative Adversarial Nets are possible solutions that could

\section{Conclusion} \label{sec:con}
In this work, we have introduced the new hardware Trojan attack on neural networks and expanded the taxonomy of neural network attacks. Several novel algorithms have been proposed to inject malicious behavior into the hardware implementation of neural networks to achieve the targeted or untargeted classification of selected input trigger. Experimental results for different adversarial scenarios have demonstrated the effectiveness of the proposed attacks. Possible defense strategies have also been discussed.

%Future work will be directed towards developing protection methods against the hardware Trojan attack on neural networks as well as advanced attacks targeting both weights and operation components or using hardware Trojan to assist other types of attacks such as adversarial examples.

\bibliographystyle{IEEEtran}
\bibliography{references}

\begin{thebibliography}{10}
\providecommand{\url}[1]{#1}
\csname url@samestyle\endcsname
\providecommand{\newblock}{\relax}
\providecommand{\bibinfo}[2]{#2}
\providecommand{\BIBentrySTDinterwordspacing}{\spaceskip=0pt\relax}
\providecommand{\BIBentryALTinterwordstretchfactor}{4}
\providecommand{\BIBentryALTinterwordspacing}{\spaceskip=\fontdimen2\font plus
\BIBentryALTinterwordstretchfactor\fontdimen3\font minus
  \fontdimen4\font\relax}
\providecommand{\BIBforeignlanguage}[2]{{%
\expandafter\ifx\csname l@#1\endcsname\relax
\typeout{** WARNING: IEEEtran.bst: No hyphenation pattern has been}%
\typeout{** loaded for the language `#1'. Using the pattern for}%
\typeout{** the default language instead.}%
\else
\language=\csname l@#1\endcsname
\fi
#2}}
\providecommand{\BIBdecl}{\relax}
\BIBdecl

\bibitem{parkhi2015deep}
O.~M. Parkhi, A.~Vedaldi, A.~Zisserman \emph{et~al.}, ``Deep face
  recognition.'' in \emph{BMVC}, vol.~1, no.~3, 2015, p.~6.

\bibitem{karu1996fingerprint}
K.~Karu and A.~K. Jain, ``Fingerprint classification,'' \emph{Pattern
  recognition}, vol.~29, no.~3, pp. 389--404, 1996.

\bibitem{chen2015deepdriving}
C.~Chen, A.~Seff, A.~Kornhauser, and J.~Xiao, ``Deepdriving: Learning
  affordance for direct perception in autonomous driving,'' in \emph{IEEE
  International Conference on Computer Vision (ICCV)}.\hskip 1em plus 0.5em
  minus 0.4em\relax IEEE, 2015, pp. 2722--2730.

\bibitem{saxe2015deep}
J.~Saxe and K.~Berlin, ``Deep neural network based malware detection using two
  dimensional binary program features,'' in \emph{10th International Conference
  on Malicious and Unwanted Software (MALWARE)}.\hskip 1em plus 0.5em minus
  0.4em\relax IEEE, 2015, pp. 11--20.

\bibitem{esteva2017dermatologist}
A.~Esteva, B.~Kuprel, R.~A. Novoa, J.~Ko, S.~M. Swetter, H.~M. Blau, and
  S.~Thrun, ``Dermatologist-level classification of skin cancer with deep
  neural networks,'' \emph{Nature}, vol. 542, no. 7639, p. 115, 2017.

\bibitem{choi2016doctor}
E.~Choi, M.~T. Bahadori, A.~Schuetz, W.~F. Stewart, and J.~Sun, ``Doctor ai:
  Predicting clinical events via recurrent neural networks,'' in \emph{Machine
  Learning for Healthcare Conference}, 2016, pp. 301--318.

\bibitem{moghaddam2016stock}
A.~H. Moghaddam, M.~H. Moghaddam, and M.~Esfandyari, ``Stock market index
  prediction using artificial neural network,'' \emph{Journal of Economics,
  Finance and Administrative Science}, vol.~21, no.~41, pp. 89--93, 2016.

\bibitem{barreno2006can}
M.~Barreno, B.~Nelson, R.~Sears, A.~D. Joseph, and J.~D. Tygar, ``Can machine
  learning be secure?'' in \emph{Proceedings of the ACM Symposium on
  Information, computer and communications security}, 2006, pp. 16--25.

\bibitem{papernot2016limitations}
N.~Papernot, P.~McDaniel, S.~Jha, M.~Fredrikson, Z.~B. Celik, and A.~Swami,
  ``The limitations of deep learning in adversarial settings,'' in
  \emph{Proceedings of the 2016 IEEE European Symposium on Security and
  Privacy}, 2016, pp. 372--387.

\bibitem{szegedy2013intriguing}
C.~Szegedy, W.~Zaremba, I.~Sutskever, J.~Bruna, D.~Erhan, I.~Goodfellow, and
  R.~Fergus, ``Intriguing properties of neural networks,'' \emph{International
  Conference on Learning Representations (ICLR)}, 2013.

\bibitem{mozaffari2015systematic}
M.~Mozaffari-Kermani, S.~Sur-Kolay, A.~Raghunathan, and N.~K. Jha, ``Systematic
  poisoning attacks on and defenses for machine learning in healthcare,''
  \emph{IEEE journal of biomedical and health informatics}, vol.~19, no.~6, pp.
  1893--1905, 2015.

\bibitem{ovtcharov2015accelerating}
K.~Ovtcharov, O.~Ruwase, J.-Y. Kim, J.~Fowers, K.~Strauss, and E.~S. Chung,
  ``Accelerating deep convolutional neural networks using specialized
  hardware,'' \emph{Microsoft Research Whitepaper}, vol.~2, no.~11, 2015.

\bibitem{chen2014dadiannao}
Y.~Chen, T.~Luo, S.~Liu, S.~Zhang, L.~He, J.~Wang, L.~Li, T.~Chen, Z.~Xu,
  N.~Sun \emph{et~al.}, ``Dadiannao: A machine-learning supercomputer,'' in
  \emph{Proceedings of the 47th Annual IEEE/ACM International Symposium on
  Microarchitecture}, 2014, pp. 609--622.

\bibitem{papernot2016towards}
N.~Papernot, P.~McDaniel, A.~Sinha, and M.~Wellman, ``Towards the science of
  security and privacy in machine learning,'' \emph{IEEE European Symposium on
  Security and Privacy}, 2016.

\bibitem{chakraborty2009hardware}
R.~S. Chakraborty, S.~Narasimhan, and S.~Bhunia, ``Hardware trojan: Threats and
  emerging solutions,'' in \emph{IEEE International High Level Design
  Validation and Test Workshop (HLDVT)}.\hskip 1em plus 0.5em minus 0.4em\relax
  IEEE, 2009, pp. 166--171.

\bibitem{tehranipoor2010survey}
M.~Tehranipoor and F.~Koushanfar, ``A survey of hardware trojan taxonomy and
  detection,'' \emph{IEEE design \& test of computers}, vol.~27, no.~1, 2010.

\bibitem{liu2017fault}
Y.~Liu, L.~Wei, B.~Luo, and Q.~Xu, ``Fault injection attack on deep neural
  network,'' in \emph{IEEE/ACM International Conference on Computer-Aided
  Design (ICCAD)}.\hskip 1em plus 0.5em minus 0.4em\relax IEEE, 2017, pp.
  131--138.

\bibitem{goodfellow2016deep}
I.~Goodfellow, Y.~Bengio, A.~Courville, and Y.~Bengio, \emph{Deep
  learning}.\hskip 1em plus 0.5em minus 0.4em\relax MIT press Cambridge, 2016,
  vol.~1.

\bibitem{akhtar2018threat}
N.~Akhtar and A.~Mian, ``Threat of adversarial attacks on deep learning in
  computer vision: A survey,'' \emph{arXiv preprint arXiv:1801.00553}, 2018.

\bibitem{moosavi2017universal}
S.-M. Moosavi-Dezfooli, A.~Fawzi, O.~Fawzi, and P.~Frossard, ``Universal
  adversarial perturbations,'' \emph{IEEE Conference on Computer Vision and
  Pattern Recognition (CVPR)}, 2017.

\bibitem{xiao2018spatially}
C.~Xiao, J.-Y. Zhu, B.~Li, W.~He, M.~Liu, and D.~Song, ``Spatially transformed
  adversarial examples,'' \emph{International Conference on Learning
  Representations (ICLR)}, 2018.

\bibitem{carlini2017towards}
N.~Carlini and D.~Wagner, ``Towards evaluating the robustness of neural
  networks,'' in \emph{IEEE Symposium on Security and Privacy (SP)}.\hskip 1em
  plus 0.5em minus 0.4em\relax IEEE, 2017, pp. 39--57.

\bibitem{tramer2017ensemble}
F.~Tram{\`e}r, A.~Kurakin, N.~Papernot, D.~Boneh, and P.~McDaniel, ``Ensemble
  adversarial training: Attacks and defenses,'' \emph{arXiv preprint
  arXiv:1705.07204}, 2017.

\bibitem{athalye2018obfuscated}
A.~Athalye, N.~Carlini, and D.~Wagner, ``Obfuscated gradients give a false
  sense of security: Circumventing defenses to adversarial examples,''
  \emph{arXiv preprint arXiv:1802.00420}, 2018.

\bibitem{papernot2016crafting}
N.~Papernot, P.~McDaniel, A.~Swami, and R.~Harang, ``Crafting adversarial input
  sequences for recurrent neural networks,'' in \emph{2016 IEEE Military
  Communications Conference}, 2016, pp. 49--54.

\bibitem{jin2008hardware}
Y.~Jin and Y.~Makris, ``Hardware trojan detection using path delay
  fingerprint,'' in \emph{2008 IEEE International Workshop onHardware-Oriented
  Security and Trust}, 2008, pp. 51--57.

\bibitem{xiao2016hardware}
K.~Xiao, D.~Forte, Y.~Jin, R.~Karri, S.~Bhunia, and M.~Tehranipoor, ``Hardware
  trojans: Lessons learned after one decade of research,'' \emph{ACM
  Transactions on Design Automation of Electronic Systems (TODAES)}, vol.~22,
  no.~1, p.~6, 2016.

\bibitem{geigel2013neural}
A.~Geigel, ``Neural network trojan,'' \emph{Journal of Computer Security},
  vol.~21, no.~2, pp. 191--232, 2013.

\bibitem{liu2017neural}
Y.~Liu, Y.~Xie, and A.~Srivastava, ``Neural trojans,'' in \emph{IEEE
  International Conference on Computer Design (ICCD)}.\hskip 1em plus 0.5em
  minus 0.4em\relax IEEE, 2017, pp. 45--48.

\bibitem{liu2017trojaning}
Y.~Liu, S.~Ma, Y.~Aafer, W.-C. Lee, J.~Zhai, W.~Wang, and X.~Zhang, ``Trojaning
  attack on neural networks,'' \emph{Department of Computer Science Technical
  Reports}, 2017.

\bibitem{tensorflow}
\BIBentryALTinterwordspacing
M.~Abadi, A.~Agarwal, P.~Barham, E.~Brevdo \emph{et~al.}, ``{TensorFlow}:
  Large-scale machine learning on heterogeneous systems,'' 2015, software
  available from tensorflow.org. [Online]. Available:
  \url{https://www.tensorflow.org/}
\BIBentrySTDinterwordspacing

\bibitem{narasimhan2012hardware}
S.~Narasimhan and S.~Bhunia, ``Hardware trojan detection,'' in
  \emph{Introduction to Hardware Security and Trust}, 2012, pp. 339--364.

\bibitem{narasimhan2013hardware}
S.~Narasimhan, D.~Du, R.~S. Chakraborty, S.~Paul, F.~G. Wolff, C.~A.
  Papachristou, K.~Roy, and S.~Bhunia, ``Hardware trojan detection by
  multiple-parameter side-channel analysis,'' \emph{IEEE Transactions on
  computers}, vol.~62, no.~11, pp. 2183--2195, 2013.

\bibitem{bhunia2013protection}
S.~Bhunia, M.~Abramovici, D.~Agrawal, P.~Bradley, M.~S. Hsiao, J.~Plusquellic,
  and M.~Tehranipoor, ``Protection against hardware trojan attacks: Towards a
  comprehensive solution,'' \emph{IEEE Design \& Test}, vol.~30, no.~3, pp.
  6--17, 2013.

\bibitem{chakraborty2009security}
R.~S. Chakraborty and S.~Bhunia, ``Security against hardware trojan through a
  novel application of design obfuscation,'' in \emph{2009 IEEE/ACM
  International Conference on Computer-Aided Design}, 2009, pp. 113--116.

\bibitem{xiao2015efficient}
K.~Xiao, D.~Forte, and M.~M. Tehranipoor, ``Efficient and secure split
  manufacturing via obfuscated built-in self-authentication,'' in \emph{IEEE
  International Symposium on Hardware Oriented Security and Trust (HOST)},
  2015, pp. 14--19.

\bibitem{goodfellow2014explaining}
I.~J. Goodfellow, J.~Shlens, and C.~Szegedy, ``Explaining and harnessing
  adversarial examples,'' \emph{International Conference on Learning
  Representations (ICLR)}, 2015.

\bibitem{samangouei2018defense}
P.~Samangouei, M.~Kabkab, and R.~Chellappa, ``Defense-gan: Protecting
  classifiers against adversarial attacks using generative models,'' \emph{6th
  International Conference on Learning Representations (ICLR)}, 2018.

\bibitem{abbasi2017robustness}
M.~Abbasi and C.~Gagn{\'e}, ``Robustness to adversarial examples through an
  ensemble of specialists,'' \emph{5th International Conference on Learning
  Representations (ICLR) Workshop}, 2017.

\end{thebibliography}

\end{document}